\documentclass{article}

\PassOptionsToPackage{numbers,sort&compress}{natbib}

\usepackage[main, final]{neurips_2026}

\usepackage[utf8]{inputenc} 
\usepackage[T1]{fontenc}    
\usepackage{url}            
\usepackage{booktabs}       
\usepackage{amsfonts}       
\usepackage{nicefrac}       
\usepackage{microtype}      
\usepackage{xcolor}         
\usepackage{placeins}

\usepackage{microtype}
\usepackage[small]{caption}
\usepackage{graphicx}
\usepackage{subcaption}
\usepackage{booktabs} 
\usepackage{natbib}
\usepackage{enumitem}

\usepackage[colorlinks=true,allcolors=blue,final]{hyperref}

\usepackage{amsmath}
\usepackage{amssymb}
\usepackage{mathtools}
\usepackage{amsthm}

\usepackage[capitalize,noabbrev]{cleveref}

\theoremstyle{plain}

\newtheorem{example}{Example}[section] 

\theoremstyle{definition}

\theoremstyle{remark}

\usepackage[textsize=tiny]{todonotes}

\title{Efficient Knowledge Transfer in Federated Bayesian Optimization through Neural Network Surrogates}

\author{%
  \textbf{Alexander Gräfe\thanks{Equal contribution.}
    \qquad Max Van Gemmeren\footnotemark[1]
  \qquad Paul Brunzema \qquad Sebastian Trimpe} \\
  Institute for Data Science in Mechanical Engineering,
  RWTH Aachen University\\
  Aachen, Germany \\
  \texttt{\{alexander.graefe, paul.brunzema, trimpe\}@dsme.rwth-aachen.de}
}

\newif\ifshowallplotlocal
\newif\ifshowallplotwarmstart
\newif\ifshowallplotnumberagents
\showallplotlocalfalse
\showallplotwarmstarttrue
\showallplotnumberagentsfalse

\usepackage{xspace}
\usepackage{amsfonts}		
\usepackage[subrefformat=parens]{subcaption}		
\usepackage{siunitx}		
\usepackage{acronym}		

\usepackage{multirow}
\usepackage{nicefrac}       
\usepackage{makecell}		
\usepackage{pifont}			

\graphicspath{{./figures/}}

\usepackage{colortbl}

\usepackage{tikz}
\usepackage[breakable]{tcolorbox}

\usepackage{booktabs}
\usepackage{siunitx}
\usepackage[table]{xcolor}   
\definecolor{bestblue}{HTML}{DCE9F7}   
\definecolor{headgray}{HTML}{F2F4F7}   
\definecolor{ourscol}{HTML}{F5F9FD}    
\usetikzlibrary{positioning}			
\usetikzlibrary{calc}					
\usetikzlibrary{arrows}
\usetikzlibrary{decorations.markings}
\usetikzlibrary{shapes}
\usetikzlibrary{spy}

\usetikzlibrary{patterns}
\usetikzlibrary{patterns.meta}
\usetikzlibrary{external}

\usepackage{ifthen}

\newboolean{symcheck}
\setboolean{symcheck}{false}

\newboolean{shownotes}
\setboolean{shownotes}{true}

\ifthenelse{\boolean{symcheck}}{
	\NewDocumentCommand{\symcheck}{m e{_^}}{%
		{\color{blue!60!black}%
			\IfNoValueTF{#2}{%
				\IfNoValueTF{#3}{#1}{#1^{\color{black}#3}}%
			}{%
				\IfNoValueTF{#3}{#1_{\color{black}#2}}{#1_{{\color{black}#2}}^{\color{black}#3}}%
		}}%
	}
}{
	\newcommand*{\symcheck}[1]{#1}
}

\newcommand{\fakepar}[1]{\textbf{#1}\quad}

\newcommand{\capt}[1]{\mdseries{\textit{#1}}}

\newcommand*{\eg}{e.g.,\xspace}
\newcommand*{\ie}{i.e.,\xspace}
\newcommand*{\cf}{cf.\xspace}

\DeclareSIUnit{\belmilliwatt}{Bm}
\DeclareSIUnit{\dBm}{\deci\belmilliwatt}

\ifthenelse{\boolean{shownotes}}
{
	\usepackage[]{todonotes}	
	\usetikzlibrary{external}

	\newcommand{\note}[4]{%
		\ifthenelse{\equal{#1}{footnote}}{{
			\tikzexternaldisable%
			\footnote{#2\todo[caption={},#3,inline]{#4}}%
		}}{{
			\tikzexternaldisable%
			\todo[caption={},#3,#1]{#4}%
		}}{}%
	}

	\newcommand{\draft}[1]{{
		\color{black!30} \vspace{1ex}\hrule\vspace{2pt}
		\color{blue!55!black} #1
		\color{black!30} \vspace{2pt}\hrule\vspace{1ex}
	}}
}
{
	\usepackage[disable]{todonotes}

	\newcommand{\note}[4]{}
	\newcommand{\draft}[1]{}
}

\acrodef{name}[\textsc{GlobaFed}]{\symcheck{Globally informed Federated Bayesian Optimization}}
\acused{name}

\acrodef{fedbo}[FedBO]{Federated Bayesian Optimization}
\acrodef{manytaskopt}[MTO]{Many-task Optimization}
\acrodef{fts}[FTS]{Federated Thompson Sampling}
\acrodef{dpftsde}[DP-FTS-DE]{Differentially Private Federated Thompson Sampling with Distributed Exploration}
\acrodef{fmtbo}[FMTBO]{Federated Many-Task Bayesian Optimization}
\acrodef{rff}[RFF]{Random Fourier Features}
\acrodef{fdemd}[FD-EMD]{Federated Data-driven Optimization with Ensemble Model Distillation}
\acrodef{rbf}[RBFN]{Radial Basis Function Network}
\acrodefplural{rbf}[RBFNs]{Radial Basis Function Networks}
\acrodef{iaffbo}[IAF-FBO]{Federated Bayesian Optimization via Optimization of an Implicit Acquisition Function}

\acrodef{bo}[BO]{Bayesian Optimization}
\acrodef{fl}[FL]{Federated Learning}
\acrodef{ml}[ML]{Machine Learning}	\acrodefindefinite{ml}{an}{a}
\acrodef{ts}[TS]{Thompson Sampling}

\acrodef{gp}[GP]{Gaussian Process}
\acrodefplural{gp}[GPs]{Gaussian Processes}
\acrodef{iid}[i.i.d.]{independent and identically distributed}
\acrodef{non-iid}[\acl{non-iid}]{non\nobreakdashes-\ac{iid}}	\acused{non-iid}

\acrodef{nn}[NN]{Neural Network}
\acrodef{mlp}[MLP]{Multi-Layer Perceptron}
\acrodefplural{mlp}[MLPs]{Multi-Layer Perceptrons}
\acrodef{vbll}[VBLL]{Variational Bayesian Last Layer}
\acrodefplural{vbll}[VBLLs]{Variational Bayesian Last Layers}
\acrodef{bll}[BLL]{Bayesian Last Layer}
\acrodef{blr}[BLR]{Bayesian Linear Regression}

\acrodef{cps}[CPS]{Cyber-Physical Systems}
\acrodef{iot}[IoT]{Internet of Things}
\acrodef{lqr}[LQR]{Linear Quadratic Regulator}

\newcommand*{\nAgents}{\ensuremath{\symcheck{N}}\xspace}
\newcommand{\setagents}{\mathcal{I}}

\newcommand{\realnmbrs}{\mathbb{R}}

\newcommand{\dataset}[1]{\mathcal{D}_{#1}}
\newcommand{\objfn}[1]{\ensuremath{J_{#1}}}
\newcommand{\numparams}[0]{\ensuremath{d}}
\newcommand{\numhidden}[0]{\ensuremath{m}}
\newcommand{\optparam}[1]{\ensuremath{x^*_{#1}}}
\newcommand{\param}[1]{\ensuremath{x_{#1}}}

\newcommand{\paramspace}[1]{\ensuremath{\mathcal{X}}}

\newcommand{\objfnappr}[1]{\ensuremath{\hat{J}_{#1}}}

\newcommand{\backboneparam}[0]{\ensuremath{\varphi}}
\newcommand{\backboneparami}[1]{\ensuremath{\varphi}_{#1}}
\newcommand{\backbone}[0]{\ensuremath{\hat{\phi}_{\backboneparam{}}}}
\newcommand{\headparam}[1]{\ensuremath{\theta_{#1}}}
\newcommand{\headf}[1]{\ensuremath{\hat{\psi}^{(#1)}_{\headparam{#1}}}}

\newcommand{\headparamrest}[1]{\ensuremath{\tilde{\theta}_{#1}}}
\newcommand{\headparamfront}[1]{\ensuremath{w_{#1}}}
\newcommand{\headparamfrontmean}[1]{\ensuremath{\bar{w}_{#1}}}
\newcommand{\headparamfrontvar}[1]{\ensuremath{S_{#1}}}
\newcommand{\headfeatures}[1]{\ensuremath{\hat{g}^{(#1)}_{\headparamrest{#1}}}}

\newcommand{\headbackbone}[1]{f^{(i)}_{\headparamrest{#1},\backboneparam{}}}

\newcommand{\plotfontsize}{8pt}
\newcommand{\benchmarktitlefontsize}{6}

\definecolor{circlegray}{HTML}{C2C2C2}

\definecolor{myblue}{HTML}{D2E1F2}
\definecolor{mygreen}{HTML}{C5E7CA}
\definecolor{myorange}{HTML}{FFF0D5}
\definecolor{mygrey}{HTML}{E5E5E5}

\tcbset{
    greybox/.style={
            colback=mygrey!25,
            colbacktitle=mygrey,
            colframe=mygrey!70!black,
            coltitle=black,
            fonttitle=\bfseries,
            boxrule=1pt,
            arc=5pt,
            left=5pt,
            right=5pt,
            top=5pt,
            breakable,
            bottom=5pt
        },
}
\newcommand{\RQ}{\hyperref[RQ]{$\textbf{RQ}$}\xspace}

\usepackage{circledsteps}
\usepackage{wrapfig}
\usepackage{multicol}

\usepackage{xcolor}
\definecolor{commentgreen}{RGB}{34,139,34}
\newcommand{\greencount}[1]{\textcolor{commentgreen}{\% #1}}
\usepackage{algorithm}
\usepackage{algorithmic}
\renewcommand{\algorithmiccomment}[1]{\greencount{#1}}

\definecolor{wongBlack}{HTML}{000000}
\definecolor{wongOrange}{HTML}{E69F00}
\definecolor{wongSkyBlue}{HTML}{56B4E9}
\definecolor{wongGreen}{HTML}{009E73}
\definecolor{wongYellow}{HTML}{F0E442}
\definecolor{wongBlue}{HTML}{0072B2}
\definecolor{wongVermillion}{HTML}{D55E00}
\definecolor{wongPurple}{HTML}{CC79A7}

\begin{document}

\maketitle

\begin{abstract}
\ac{fedbo} has emerged as a powerful paradigm for enhancing \ac{bo} in distributed, privacy-sensitive settings, enabling agents to accelerate the optimization of their local objectives by sharing knowledge about their similarities without exposing raw data.
However, existing methods rely on restrictive assumptions, such as near-identical objective functions or shared Gaussian process hyperparameters that are often violated in heterogeneous real-world applications.
We present \ac{name}, a novel framework that enables efficient knowledge transfer across different but related objectives without compromising privacy.
At its core is a probabilistic neural network surrogate with a shared feature-extracting backbone and agent-specific heads equipped with variational Bayesian last layers 
to provide uncertainty estimates.
During optimization, agents use federated learning on the shared backbone to collaboratively build a common latent feature representation, while agent-specific heads adapt to their individual objectives.
Our empirical results demonstrate that \ac{name} significantly accelerates convergence and reaches better final performance compared to state-of-the-art \ac{fedbo} baselines, and enables agents to warm-start their optimization from previously learned global features.
\end{abstract}

\section{Introduction}
\label{sec:introduction}

\acf{bo} has become the de facto standard for the sample-efficient tuning of black-box systems where evaluations are expensive~\cite{garnett_bayesoptbook_2023}.
In real-world deployments, ranging from healthcare devices and wearable exoskeletons to autonomous vehicles, \ac{bo} enables populations of deployed units to adapt to dynamic environments and heterogeneous user-defined objectives~\cite{kim2019bayesian, lu2021bayesian,chakrabarty2021scalable,lin2023bayesian, menn2024lipschitz,phatale2024enhancing, kim2025pilot}. 
Crucially, while each individual unit (e.g., a specific exoskeleton) optimizes its own unique objective function, these functions remain inherently related
across the fleet because the underlying systems share the same physical characteristics or structural dynamics.
However, traditional \ac{bo} methodologies are predominantly designed for single-instance optimization, treating each system as an isolated agent.
This ``isolated agent'' paradigm forces every new agent to undergo an individual tuning process from scratch, which can be regarded as a cold-start penalty in situations where other agents (and thus data) are available \emph{in principle}. 
In domains where every sample represents a physical experiment or a human-in-the-loop interaction, failing to exploit these cross-agent relationships introduces inefficiencies.

A fundamental challenge in leveraging these relationships is ensuring 
privacy protection.
Because many applications involve sensitive personal data, like individual health metrics or driving patterns, knowledge sharing must occur without compromising local information.
\acf{fedbo} has emerged to address these privacy concerns and facilitate rapid distributed optimization through inter-agent knowledge transfer~\cite{dai2020fts,dai2021ftsdiff,zhu2023fmtbo,wang2024fdemd,liu2025optimization}.

However, current \ac{fedbo} approaches often rely on overly strong assumptions about cross-agent similarity.
In particular, they presume near-identical objective functions~\cite{dai2020fts,dai2021ftsdiff} 
or closely matched hyperparameters of the underlying \ac{gp} models \cite{zhu2023fmtbo}.
Such assumptions substantially limit their applicability in real-world settings, where heterogeneity across agents is the norm rather than the exception.
For instance, differing user preferences (\eg{} ``sporty” versus ``comfort” driving styles) or variations in physical scale (\eg{} exoskeletons fitted to users of different body sizes) can induce objective functions that differ markedly in both magnitude and lengthscale.
As a result, existing \ac{fedbo} methods struggle to accommodate agents that are distinct yet share underlying structural characteristics.
These limitations motivate this paper's guiding research question.

\textbf{Guiding Research Question (RQ):} How can we design a \ac{fedbo} method that delivers \textit{(i)} fast convergence with strong final performance, \textit{(ii)} without necessitating strong assumptions, \textit{(iii)} while ensuring privacy?

\subsection{Approach Overview}\label{sec:introduction:overview}

\begin{wrapfigure}{r}{0.5\textwidth}
    \centering
    \vspace{-1em}
    \includegraphics[width=0.5\textwidth]{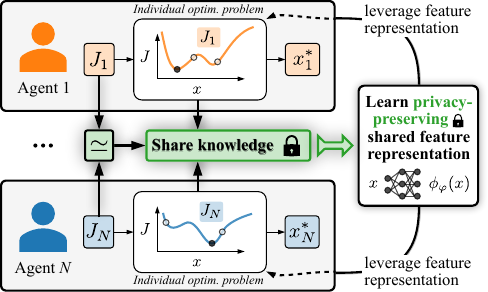}%
    \caption{
    \ac{name}. 
    \capt{$K$ agents have different but related objective functions.
    \acs{name} leverages federated learning combined with \acs{nn} surrogate models for \acs{bo} to learn a shared feature representation for efficient knowledge transfer while preserving privacy, resulting in faster convergence on the individual tasks of each agent.}%
    }%
    \vspace{-1em}
    \label{fig:overview}
\end{wrapfigure}

To address the aforementioned research question, this work introduces \acf{name}
(\cref{fig:overview}).
At the core of \ac{name} is a probabilistic \ac{nn} surrogate model, specifically a \ac{vbll} model~\cite{harrison2024variational,brunzema2025bovbll}, consisting of a \textit{shared backbone} and \textit{agent-exclusive heads}  
that overcome the limitations of rigid \ac{gp} priors (\RQ~\textit{(ii)}). 
Knowledge transfer is facilitated through personalized \acf{fl}~\cite{arivazhagan2019federated,collins2021exploiting,pillutla2022federated}.
The backbone learns a common feature representation of the underlying system behavior across all agents, while the local heads adapt to individual objective functions.
To achieve this, agents train the full model on local data and exchange backbone weights for global aggregation, ensuring that sensitive raw data is never shared (\RQ~\textit{(iii)}).
Within each \ac{bo} iteration, agents utilize the aggregated backbone and their local heads to construct individual acquisition functions that are informed by collective insights.
Through the knowledge transfer via a shared latent space, \ac{name} enables agents to achieve significantly faster convergence and strong final performance (\RQ~\textit{(i)}).


\fakepar{Contributions}
In summary, our contributions are:
\begin{enumerate}[topsep=0pt] \setlength{\itemsep}{0pt} \setlength{\parskip}{0pt} \setlength{\parsep}{0pt}
    \item[\textbf{C1}]
    A novel \ac{fedbo} framework, termed \Ac{name}, that accelerates optimization via privacy-preserving knowledge sharing while relying on weaker task similarity requirements than existing methods.
    \item[\textbf{C2}] A federated surrogate-training scheme that collaboratively learns a shared neural feature representation through backbone aggregation, while keeping each agent’s observations and \ac{vbll} heads local for task-specific Bayesian uncertainty quantification. 
    \item[\textbf{C3}] 
    Comprehensive experiments showing that \ac{name} exhibits considerable performance improvements both in convergence speed and final performance compared to existing state-of-the-art baselines when objectives are related but structurally different, performs on par on a heterogeneous Many-Task benchmark, and accelerates convergence for late-joining agents through warm-started shared features.
\end{enumerate}

\section{Problem Formulation}\label{sec:problem-formulation}

This section formalizes our \RQ.
We formalize our problem following the standard \ac{manytaskopt} setup~\cite{dai2020fts,dai2021ftsdiff,zhu2023fmtbo,wang2024fdemd,liu2025optimization}.
We consider a set $\setagents{}$ consisting of $\nAgents{}$ agents. 
Each agent $i \in \setagents{}$ aims to identify a set of parameters $\optparam{i} \in \paramspace{} \subseteq \realnmbrs{}^\numparams{}$ that maximizes its individual objective function $\objfn{i}: \realnmbrs{}^\numparams{}\to\realnmbrs{}$:
\begin{equation}\label{eq:problem}
    \forall i\in\setagents{}:~\optparam{i} = \arg\max_{\param{}\in\paramspace{}}\objfn{i}(\param{}).
\end{equation}

\fakepar{Observation Model}
We treat the objective functions $\objfn{i}$ as black-box models, where agents lack knowledge of the analytical expressions. 
At each round $t\in \{1, \dots, T\}$, where $T$ is the query budget, every agent $i$ independently selects a query point $x_t^{(i)}$ and receives noisy zeroth-order feedback
\begin{equation}
    \label{eq:query}
    y_t^{(i)} = \objfn{i}\left(\param{t}^{(i)}\right) + \epsilon_{t}^{(i)}, \quad \epsilon_{t}^{(i)} \sim \mathcal{N}(0,\sigma^2_i),
\end{equation}
where evaluations are corrupted by \ac{iid} Gaussian noise with agent-specific variance $\sigma^2_i$.
The functions have high sampling cost stating that such queries are expensive (\eg{} obtained through a physical experiments).

\fakepar{Privacy Setting} 
We assume that raw samples $(\param{t}^{(i)}, y_t^{(i)})$ contain privacy-sensitive information and must remain \textit{agent-local}, \ie{} they are never shared between agents. 
We focus exclusively on this aspect of data privacy to demonstrate that our approach to privacy-preserving knowledge transfer in \ac{fedbo} via modern probabilistic \ac{nn} surrogates works in principle. 
Consequently, we view defenses against other equally relevant privacy concerns, such as gradient inversion attacks~\cite{zhu2019deep, huang2021evaluating, hatamizadeh2023gradient, geng2023improved, zhou2025adversarial}, as largely orthogonal to our core contribution. Indeed, our framework can readily incorporate standard defenses against such attacks, including gradient perturbation~\cite{zhu2019deep}, input mixing~\cite{huang2020instahide}, or representation perturbation~\cite{sun2021soteria}.


\fakepar{Task Similarity}
The objectives are related, meaning that samples from one function yield information about the others. 
Rather than assuming a fixed relational structure, we aim to develop a unified framework capable of capturing diverse scenarios, such as:
\begin{itemize}[leftmargin=*, noitemsep, topsep=0pt]
    \item \textit{Near-Identical Objectives.} The objectives for any pair of agents $i,j$ satisfy $\objfn{i}(\param{}) \approx \objfn{j}(\param{})$, exhibiting only minor deviations arising from, \eg{} manufacturing tolerances.
    \item \textit{Linear Combinations of Basis Functions.} Objectives are formulated as $\objfn{i}(\param{}) = \sum_{b=1}^{B} c_{b}^{(i)}f_{b}(\param{})$. The coefficients $c_{i,\mathrm{b}}$ vary among agents, modeling user preferences (e.g., balancing comfort vs.\ energy efficiency).
    \item \textit{Parametric System Variations.} Objectives arise from identical system types operating under different physical parameters (\eg{}
    exoskeletons for different body sizes).
\end{itemize}


\fakepar{Problem Statement} 
The goal of this paper is to solve \eqref{eq:problem} for all agents by maximizing sample efficiency, leveraging relations between objective functions without assuming their specific type, and strictly adhering to privacy constraints by keeping raw samples agent-local to address \RQ.

\section{Related Work}\label{sec:related-work}
This section reviews \ac{fedbo} methods for the \ac{manytaskopt} problem described above.
All methods follow the standard \ac{bo} loop~\cite{garnett_bayesoptbook_2023}:
they train a surrogate model, select the next query point, and evaluate the objective.
The main difference is how knowledge is transferred between agents, which determines the supported type of relationships between objective functions.


\Ac{fts} and its differentially private extension \ac{dpftsde}~\cite{dai2020fts,dai2021ftsdiff} let each agent fit a \ac{gp} and approximate the posterior via \ac{rff}, which allows them to share their \ac{gp} without revealing private data.
Agents randomly choose between using their own \ac{gp} and another agent's \ac{rff}-weights when sampling.
Consequently, an agent's \ac{gp} must be directly informative for others, and \ac{fts} effectively requires \emph{closely aligned objective functions}.

\ac{fmtbo}~\cite{zhu2023fmtbo} instead shares \ac{gp} hyperparameters, such as the length scale, rather than the \ac{gp} itself. 
These hyperparameters are averaged using similarity-based weights. 
Each agent keeps a local \ac{gp} (using only local data) and a global \ac{gp} (using aggregated hyperparameters) to choose the next query. 
Thus, the relationship assumed by \ac{fmtbo} is that \emph{objective functions are generated by \acp{gp} with similar hyperparameters}.

Instead of \acp{gp}, \ac{fdemd}~\cite{wang2024fdemd} employs \acp{rbf} as surrogate models. 
Agents share weights with a server, which constructs an ensemble \ac{rbf} and distills a student network.
Agents then combine this network with their local \ac{rbf} to select query points. 
\Ac{fdemd}'s knowledge transfer therefore \emph{requires that objectives are closely aligned}.

In \ac{iaffbo}~\cite{liu2025optimization}, agents maintain a local \ac{gp} and distill it into a classifier that predicts which of two points has a higher acquisition value. 
These classifiers are shared and clustered 
based on weight similarity. 
Each cluster's weights are averaged, and the resulting classifiers guide the next queries. 
This mechanism hence implicitly assumes that \emph{objectives are closely aligned within clusters}.

To summarize, existing \ac{fedbo} methods rely on restrictive similarity assumptions. 
They either require closely aligned objectives~\cite{dai2020fts,dai2021ftsdiff,wang2024fdemd,liu2025optimization}, which is already violated by simple linear combinations of basis functions (cf. \cref{sec:problem-formulation}), or they assume similar underlying \ac{gp} hyperparameters~\cite{zhu2023fmtbo}, which basic input scaling, \eg{} when one agent optimizes a fast system and another a slow one, breaks. 
In contrast, \ac{name} removes these constraints and does not rely on restrictive assumptions. 
We achieve this via probabilistic \ac{nn} surrogate models with a shared backbone and exclusive heads, which has not been proposed for \ac{fedbo} before.
This architecture facilitates a more flexible representation of objective function relations.

\fakepar{Methods related to \ac{manytaskopt}}
While several frameworks address multi-task optimization, they differ significantly from our setting in their assumptions regarding data locality and task relationships.
Standard multi-task \ac{gp} and \ac{bo} approaches~\cite{swersky2013multi} typically rely on an explicit task-relation kernel to model dependencies between tasks.
However, performing inference in these models requires centralized access to the data from all tasks, which is infeasible in the federated setting we consider.
Similarly, contextual \ac{gp} and \ac{bo} methods~\cite{krause2011contextual} assume that tasks can be characterized by additional task-specific context or features.
Beyond the challenge of defining such context in many practical applications, these methods still require a central entity to aggregate data from all agents to train the global surrogate, thereby violating the privacy constraints inherent in \RQ.
In contrast, \ac{name} facilitates knowledge transfer without requiring centralized data access or predefined task descriptors.

Perrone et al.~\cite{perrone2018scalable} propose a multi-task transfer learning approach for \ac{bo}, combining task-specific Bayesian linear regression surrogates with a shared neural feature map. 
Both are learned jointly by optimizing a sum of task-specific marginal likelihoods, and the model can be retrained as new \ac{bo} observations become available. 
The approach demonstrates transfer by using evaluations from previous hyperparameter optimization tasks to warm-start a new target task and by exploiting related auxiliary signals, such as training loss or evaluation cost. 
Similarly, \ac{name} combines a shared representation with task-specific Bayesian heads. 
However, unlike \cite{perrone2018scalable}, agents optimize concurrently using local observations and retain private \ac{vbll} heads, exchanging only backbone parameters for federated aggregation without pooling task data.


\section{Globally Informed Federated BO}

\begin{figure}
    \centering
    \includegraphics[width=1.0\textwidth]{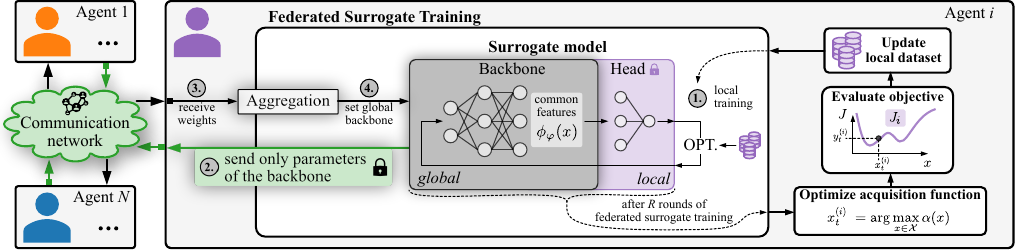}
    \caption{
        Overview of \ac{name}. 
        \textit{In each iteration, agents} 1. 
        \textit{train their surrogate models on local data,} 2. \textit{broadcast their learned backbone parameters,} 3. \textit{receive weights from other agents, and} 4. \textit{aggregate these weights and use them as initialization of the backbone for the next round.
        Through this, privacy-sensitive samples of the objective function remain agent-local. 
        }
        }%
    \label{fig:method}
    \vspace{-1em}
\end{figure}

We present our method addressing the stated problem. 
\Cref{sec:method:surrogate} first discusses our proposal to leverage probabilistic \ac{nn} surrogates and specifically \ac{vbll} models, as opposed to \acp{gp}, which a majority of \ac{fedbo} approaches rely on.
\Cref{sec:method:training} then details the training of these surrogates to achieve efficient knowledge transfer between agents while ensuring privacy.
Lastly, in \cref{sec:method:operation}, we discuss the overall operation of \ac{name}.

\subsection{Decomposed \ac{nn} Surrogates for Flexible Modeling}
\label{sec:method:surrogate}

\ac{name} uses a decomposed surrogate architecture that balances global knowledge transfer with agent-specific specialization.
For each agent, the surrogate model is
\begin{equation}\label{eq:modeling}
    \objfnappr{i}(\param{}) = \headf{i}(\backbone{}(\param{})),
\end{equation}
where $\backbone{}:\realnmbrs{}^\numparams{}\to\realnmbrs{}^\numhidden{}$ is a shared \ac{mlp} backbone with parameters $\backboneparam{}$, and $\headf{i}:\realnmbrs{}^\numhidden{}\to\realnmbrs{}$ is an agent-specific head with parameters $\headparam{i}$.
This structure allows \ac{name} to model various degrees of relationships between objectives, as illustrated by the following examples:

\begin{example}[Equal Objectives]
    Consider tasks where $\objfn{i} = \objfn{j}$ for all $i,j$.
    In this setting, the surrogate admits an optimal solution in which all heads coincide, \ie{} $\headparam{i} = \headparam{*}$ for all $i$, meaning all surrogate models are equal and approximate the same function. 
\end{example}

\begin{example}[Linear Combination of Basis Functions]
    Suppose that the objectives are linear combinations of shared basis functions (\cf{} \cref{sec:problem-formulation}).
    In this case, the backbone would learn the shared basis functions and the heads the agent-specific linear transformation.
\end{example}

\begin{example}[Weakly Related or Unrelated Functions]
    For unrelated objectives, \eqref{eq:modeling} avoids negative transfer by allowing the shared backbone to learn a task-agnostic mapping. 
    By setting $\numhidden{} = \numparams{}$, the backbone may reduce to an identity mapping,
    $
        \backbone{}(\param{}) \approx \param{}.
    $
    If the local head $\headf{i}$ has sufficient expressive capacity, the surrogate becomes
        $\objfnappr{i}(\param{})
        = \headf{i}\!(\backbone{}(\param{}))
        \approx \headf{i}(\param{})
        \approx \objfn{i}(\param{})$.
\end{example}


To explicitly model predictive uncertainty, we adopt \ac{vbll} models~\cite{harrison2024variational,brunzema2025bovbll} as the final layer of the surrogate's head.
\ac{vbll} models are structurally similar to standard \acp{mlp}, but replace the deterministic last linear layer with a \ac{blr} layer.
Unlike other Bayesian last layer approaches that rely on explicit conditioning to infer the posterior over the final-layer weights~\cite{snoek2015scalable,watson2021latent,fiedler2023improved}, \acsp{vbll} directly learn the parameters of a variational posterior $q_{i}(\headparamfront{i}) = \mathcal{N}(\headparamfrontmean{i},\headparamfrontvar{i})$.
Given this variational posterior and the observation model in \eqref{eq:query}, conjugacy yields a tractable posterior predictive distribution for each agent's surrogate model of the individual objective:
\begin{align}
    \objfnappr{i}(\param{}) \sim\mathcal{N}(\headparamfrontmean{i}^\top\headbackbone{i}(\param{}),\headbackbone{i}(\param{})^\top\headparamfrontvar{i}\headbackbone{i}(\param{}) + \sigma_i^2),
\end{align}
where $\headbackbone{i}(\param{}) = \headfeatures{i}(\backbone(\param{}))$ and $\headfeatures{i}(\param{})$ comprises all head layers except the last one with parameters \headparamrest{i}.
Hence, the learnable head parameters are $\headparam{i} = \{\headparamrest{i}, \headparamfrontmean{i}, \headparamfrontvar{i}, \sigma_i^2 \}$.

\subsection{Surrogate Model Training Ensuring Privacy and Knowledge Transfer}\label{sec:method:training}
The effectiveness of \ac{name} relies on the ability of the surrogate models to capitalize on shared structural similarities across agents without compromising the privacy of local observations.
Having defined the decomposed surrogate architecture in \cref{sec:method:surrogate}, we now describe a tailored \ac{fl} procedure that enables efficient global knowledge transfer while ensuring privacy.

\fakepar{Local Surrogate Training}
Each agent trains a surrogate model on its local dataset $\dataset{i} = \{(\param{t}^{(i)}, y_t^{(i)})\}_{t=1}^T$, which consists of noisy evaluations (\cf{}~\eqref{eq:query}).
Training is performed by maximizing the variational lower bound on the marginal likelihood
\cite{harrison2024variational,brunzema2025bovbll}, given by
\begin{align}
    \label{eq:variational_lower_bound}
    \mathcal{L}(\headparam{i},\backboneparam{})
    &\coloneqq\!
    \frac{1}{T}\sum\limits_{t=1}^T
    \left[
    \log\mathcal{N}\!\left(
    y_t \mid
    \headparamfrontmean{i}^{\top}\headbackbone{i}(\param{}),
    \sigma_i^2
    \right)
    -\frac{1}{2}
    \headbackbone{i}(\param{})^{\top}
    \headparamfrontvar{i}
    \headbackbone{i}(\param{})
    \sigma_i^{-2}
    \right]
    -
    \mathcal{R}(\headparam{i},\backboneparam{}),
    \\
    &\text{where}\quad
    \mathcal{R}(\headparam{i},\backboneparam{})
    \coloneqq
    \frac{1}{T}
    \Big[
    \mathrm{KL}\!\left(
    q_i(\headparamfront{i})
    \,\middle\|\,
    p(\headparamfront{i})
    \right)
    -\log p \left(
    \{ \headparamrest{i}, \backboneparam{}\}
    \right)
    -\log p\!\left(
    \sigma_i^2
    \right)
    \Big].
    \nonumber
\end{align}
where $\text{KL}(q_{i}(\headparamfront{i})|| p(\headparamfront{i}))$ is the Kullback-Leibler divergence between the variational posterior and the prior, and negative log-priors imposed on $\{\headparamrest{i},\backboneparam{}\}$ and $\sigma_i^2$.
In practice, additional regularization terms are applied to the backbone and the noise. 
\begin{wrapfigure}{r}{0.5\textwidth}
\vspace{-0.5em}
\begin{minipage}{0.48\textwidth}
\begin{algorithm}[H]
   \caption{Federated Surrogate Training on Agent $i$}
   \label{alg:aprroach_fl_surrogate_training}
\begin{algorithmic}[1]
    \footnotesize
   \STATE {\bfseries Input:} Local dataset $\dataset{i}$, Rounds $R$
   \STATE {\bfseries Initialize:} $\headparam{i}[0]$, $\backboneparam[0]$ \hfill \COMMENT{Random initialization}
   
   \FUNCTION{FedSurrogateTraining[$\dataset{i}$]}
      \FOR{$r=0$ {\bfseries to} $R-1$}
         \STATE \COMMENT{Local surrogate training maximizing \eqref{eq:variational_lower_bound}}
         \STATE $(\headparam{i}[r+\tfrac{1}{2}], \backboneparami{i}[r+\tfrac{1}{2}]) \gets 
         \text{Train}([\headparam{i}[r], \backboneparam[r]], \dataset{i})$
         
         \STATE \COMMENT{Communication \& aggregation}
         \STATE Send $\backboneparami{i}[r+\tfrac{1}{2}]$ to communication network
         \STATE $\Omega \gets \{ \backboneparami{j}[r+\tfrac{1}{2}] \mid \forall j \in \setagents \}$ $\gets \text{Receive()}$
         \STATE $\backboneparam[r+1] \gets \text{AggregateBackbones}(\Omega, \backboneparam[r])$
         
         \STATE \COMMENT{Surrogate head finetuning}
         \STATE $\headparam{i}[r+1] \gets 
         \text{TrainHead}([\headparam{i}[r+\tfrac{1}{2}], \backboneparam[r+1]], \dataset{i})$
      \ENDFOR
      \STATE {\bfseries return} $\headparam{i}[R], \backboneparam[R]$
   \ENDFUNCTION
\end{algorithmic}
\end{algorithm}
\end{minipage}
\vspace{-1em}
\end{wrapfigure}
We refer to~\citet[Algorithm~1]{harrison2024variational} for the full formulation as well as the details on the parameterization of $q_i$ through Cholesky factors of $S_i$.
The objective in \eqref{eq:variational_lower_bound} jointly optimizes the variational posterior together with the remaining head parameters $\headparamrest{i}$ and the backbone parameters $\backboneparam$.
All optimization is performed locally, and no raw data are shared between agents.

\fakepar{Federated Knowledge Transfer}
To enable knowledge transfer across agents, we embed the training procedure in a personalized \ac{fl} loop~\cite{arivazhagan2019federated,collins2021exploiting,pillutla2022federated} shown in~\cref{alg:aprroach_fl_surrogate_training}.
After randomly initializing the head and backbone parameters, agents perform $R$ communication rounds.
In each round, agents update their local models by optimizing
\eqref{eq:variational_lower_bound} on their private datasets.
Subsequently, only the backbone parameters are shared and aggregated using an \ac{fl} method.
In our implementation, we employ FedOpt~\cite{reddi2021adaptive}, which treats local weight updates as pseudo-gradients and aggregates them using an ADAM optimizer to account for heterogenous data distributions caused by different objectives.
The resulting aggregated backbone defines a shared feature representation for the next round.
The end of the loop is always a fine-tuning of the head parameters given the aggregated backbone.



\subsection{Overall \Ac{name} Operation Through Agent-local Thompson Sampling}
\label{sec:method:operation}


\begin{wrapfigure}{r}{0.53\textwidth}
\vspace{-2.5em}
\begin{minipage}{0.48\textwidth}
\begin{algorithm}[H]
\caption{BO Loop of Agent $i$}
\label{alg:agentoperation}
\begin{algorithmic}[1]
\footnotesize
    \STATE $\dataset{i} \gets $ Initial set of observations
    \FOR{$t= |\dataset{i}|$ {\bfseries to} $T$}
        \STATE $(\headparam{i}[t], \backboneparam[t])$ $\gets$ FedSurrogateTraining($\dataset{i}$)
        \STATE Build local surrogate $\objfnappr{i}(\param{})$ from $\headparam{i}[t]$ and $\backboneparam[t]$
        \STATE $\param{t}^{(i)} \gets \arg\max_{\param{}\in\mathcal{X}} \alpha (\objfnappr{i}(\param{}), \dataset{i})$ 
        \STATE $y_t^{(i)} \gets$ QueryObjective($\param{t}^{(i)} $)
        \STATE $\dataset{i} \gets \dataset{i} \cup \{(\param{t}^{(i)} , y_t^{(i)})\}$
    \ENDFOR
    \STATE \textbf{return} $\arg\max_{(\param{}, y) \in \dataset{i}} y$
\end{algorithmic}
\end{algorithm}
\end{minipage}
\vspace{-1em}
\end{wrapfigure}

The overall agent operation
follows the standard \ac{bo} framework for \ac{vbll} models as surrogates~\cite{brunzema2025bovbll} and is shown in \cref{fig:overview} and \cref{alg:agentoperation}.
At every iteration $t$, the agents use their obtained surrogate models from the previous section to build an acquisition function $\alpha (\objfnappr{i}(\param{}), \dataset{i})$ which is globally maximized for the next query location.
While \ac{vbll} models allow for a variety of acquisition functions, we adopt Thompson sampling as a convenient acquisition function for such \ac{nn} surrogate models.
For this, we first sample a realization of the last layer weight distribution $w^{(i)}_{t}\sim q_i(\headparamfront{i})$. 
The Thompson sample is then constructed as a linear model using the sample from the last layer as $\alpha_{TS}(\param{}) = {(w^{(i)}_{t})}^\top\headfeatures{i}(\backbone(\param{}))$.
It is then maximized using ten multi-starts with L-BFGS-B \citep{byrd1995limited} leading to the next query point $\param{t}^{(i)}$.
Each agent $i$ then queries the objective function at their respective new query locations to get a noisy measurement and updates its dataset $\dataset{i}$.
The agents then continue with the next iteration and train the surrogate models on the new datasets.

\section{Experiments}\label{sec:experiments}
In this section, we demonstrate that our approach accelerates the optimization process in \ac{manytaskopt} problems through efficient knowledge transfer\footnote{Code is available at
\url{https://github.com/Data-Science-in-Mechanical-Engineering/GlobaFed}.}.
Our main results are:
\begin{itemize}[leftmargin=*, noitemsep, topsep=0pt]
    \item \Ac{name} substantially outperforms existing \ac{fedbo} methods and agent-local \ac{bo}, which does not leverage knowledge transfer, under the same optimization budget.
    \item \Ac{name} allows agents that join later to immediately exploit shared features encoding information from earlier agents for effective warm-starts.
    \item \Ac{name} is robust to negative knowledge transfer achieving comparable performance to existing \ac{fedbo} methods on an \ac{manytaskopt} benchmark with multiple task groups.
\end{itemize}

\subsection{Experimental Setup}\label{sec:experiments:setup}

\fakepar{Benchmark Problems}
We consider two sets of benchmark problems, which we summarize here and detail in Appendix~\ref{sec:appendix:benchmark}. 
The first benchmark, referred to as the related-task benchmark (\cref{sec:experiments:homogeneous-benchmark}), comprises nine tasks derived from objective functions that are systematically transformed to yield distinct yet related objectives for each agent.
The first five tasks are based on standard benchmark functions in which both the input and the output value (\ie the location and height of the optimum) are shifted.
These include the Ackley function (in 1D, 5D, and 10D), the Branin function (2D) and the Hartmann function (6D).
The \emph{combination} (Comb.) task is a linear combination of five base functions (\cf \cref{sec:method:surrogate}) with 1, 5 and 10 input dimensions. 
The last task is a controller tuning problem (5D) as a real-world example, where the agents tune controllers for cart-pole systems that differ in their physical parameters.
Figures with selected benchmarks report Ackley (1D), cart-pole and Comb. (10D) as representative examples; full results for the remaining functions are included either in the main figures or in \cref{sec:appendix:extended-results}.

The second benchmark is the many-task benchmark~\cite{da2017evolutionary} (\cref{sec:experiments:manytask}). 
It comprises 18 distinct 10-dimensional objective functions that, unlike the related-task benchmark, may not share common structure (\cf \cref{sec:appendix:benchmark}).
While this benchmark does not reflect the primary use case of \ac{name}, which targets settings with shared task structure, it provides a valuable testbed for evaluating robustness to potential negative knowledge transfer in more heterogeneous scenarios.


\fakepar{Baselines}
We compare our approach against baselines with different levels of information sharing. 
First, as alternative \ac{fedbo} baselines, we compare our method against \ac{dpftsde}~\cite{dai2021ftsdiff}, \ac{fmtbo}~\cite{zhu2023fmtbo}, \ac{fdemd}~\cite{wang2024fdemd} and \ac{iaffbo}~\cite{liu2025optimization}.
Second, we include \emph{agent-local \ac{bo}} with either \ac{gp} surrogate model (L-GP) or a \ac{vbll} surrogate (L-VBLL) to compare \ac{name} with methods without any knowledge sharing.
Third, to assess the influence of privacy constraints on performance, we include a \emph{centralized version} of our approach, where we train the surrogate model on the entire dataset, \ie{} with data sharing. The full details on its training and results can be found in Appendix~\ref{sec:appendix:central-baselines}.



\fakepar{Hyperparameters}
If not specified otherwise, we perform all experiments with 10 agents and $\numparams{}$ initial points.
To evaluate the generalization capabilities, we tuned the hyperparameters, which can be found in Appendix~\ref{sec:appendix:ablation}, for all \ac{nn}-based approaches once on the shifted 10-dimensional Ackley problem from the related-task benchmark. 
These hyperparameters were then kept fixed for all other benchmark problems to test whether a single configuration performs robustly across tasks. 

Beyond our main results, we conduct a comprehensive ablation study on key hyperparameters to quantify their impact on performance. 
In this section, we highlight the effect of varying the number of agents, while the remaining ablation results are deferred to Appendix~\ref{sec:appendix:ablation}.


\fakepar{Metrics}
We measure each agent's simple regret at iteration $t$ to evaluate how quickly each approach identifies high-performing parameters. 
Results for the related-task benchmark report the average simple regret across agents, since the magnitudes of objectives are comparable.
For the many-task benchmark, where objectives' magnitudes differ substantially and regret scales vary, we report the simple regret for each agent individually.


\subsection{Related-Task Benchmark}\label{sec:experiments:homogeneous-benchmark}
This section evaluates \ac{name} on the related-task benchmark, comparing it with related \ac{fedbo} methods and agent-local baselines. 
We first consider a standard setting where all agents start simultaneously, and then a warm-start setting where one agent joins the optimization later.

\begin{figure}[t]
    \centering
    \input{plots/homogeneous_benchmark_fedbo_all_wide}
    \vspace{-1em}
\end{figure}

\Cref{fig:experiments:homogeneous-fedbo} compares \ac{name} with related \ac{fedbo} approaches. 
In low dimensions, all methods except \ac{iaffbo} and \ac{fdemd} quickly reach low regret, whereas in higher dimensions \ac{name} clearly dominates, converging faster and achieving the lowest final regret of all methods across all benchmarks, including rapid convergence on the controller optimization task.
On Combination, \ac{fmtbo} attains lower regret initially, yet \ac{name} overtakes it after a few iterations.
These results show that the \ac{nn} surrogate learns expressive shared features that the head can adapt to guide the search.
As a result, \ac{name} enables effective knowledge transfer, particularly in high dimensions.
In contrast, the restrictive assumptions of \ac{dpftsde}, \ac{fmtbo}, \ac{iaffbo} and \ac{fdemd} (cf. \cref{sec:related-work}) limit information sharing about task relations, which slows down optimization. 

\fakepar{Comparison with agent-local baselines}
We compare \ac{name} with agent-local approaches in \cref{fig:experiments:homogeneous-local}. 
For low-dimensional problems, our method matches agent-local L-GP and outperforms agent-local L-VBLL. 
As dimensionality increases, \ac{name} substantially outperforms all agent-local variants, showing faster convergence and lower final regret.
\ifshowallplotlocal
\begin{figure}[t]
    \centering
    \input{plots/homogeneous_benchmark_local_extended.tex}
    \vspace{-1em}
\end{figure}
\else
\begin{wrapfigure}{r}{0.49\textwidth}
    \vspace{-1em}
    \centering
    \input{plots/homogeneous_benchmark_local.tex}
    \vspace{-2em}
\end{wrapfigure}
\fi

\fakepar{Comparison with \ac{fedbo} baselines}
These results suggest that simple low-dimensional problems are well handled by traditional \ac{bo} with \acp{gp} and require no knowledge transfer. 
In high dimensions, however, knowledge transfer significantly speeds up convergence.
Since agents sample different initial points, they explore different regions of the search space and collectively gather more information than purely agent-local methods. 

\fakepar{Warmstarting}\label{sec:experiments:homogeneous-benchmark:warmstart}
Finally, we study a scenario where an agent joins the optimization at a later iteration and leverages the collective knowledge of existing agents to accelerate its search. We refer to late-joining agents as warm-started and to those present from the beginning as cold-started.
Our experiments compare the warm-starting performance of \ac{name}, \ac{fmtbo}, \ac{dpftsde}, \ac{iaffbo} and \ac{fdemd}.


\Cref{fig:experiments:homogeneous-fedbo-warmstart} shows our results. 
Warm- and cold-started agents exhibit similar initial regret. 
In low-dimensional problems, all methods reduce regret at comparable rates compared to the cold-started agents, yielding no notable warm-start advantage. 
In high-dimensional problems, however, \ac{name} enables warm-started agents to match the regret of cold-started agents within only a few iterations, whereas in \ac{fmtbo}, \ac{dpftsde}, \ac{iaffbo} and \ac{fdemd}, warm-started agents improve at roughly the same pace as the cold-start agents. 

\ifshowallplotwarmstart
\begin{figure}[t]
    \centering    
    \input{plots/homogeneous_benchmark_fedbo_warmstart_extended.tex}
    \vspace{-1em}
\end{figure}
\else
\begin{wrapfigure}{r}{0.49\textwidth}
    \centering
    \vspace{-0.2em}
    \centering
    \input{plots/homogeneous_benchmark_fedbo_warmstart.tex}
    \vspace{-3em}
\end{wrapfigure}
\fi
These experiments demonstrate that the shared features learned by \ac{name} encode global knowledge that generalizes to unseen objectives. 
This efficient knowledge transfer allows warm-started agents to train accurate local heads with just a few observations and accelerate optimization.

\subsection{Many-Task Benchmark}\label{sec:experiments:manytask}

This section presents the results of our approach on the many-task benchmark shown in \cref{tab:experiments:manytask_10d}. 
We use this benchmark to assess how well our algorithm performs in a setting where only a subset of tasks is related.
%
%
Our approach achieves results comparable to recent \ac{fedbo} methods.
Generally, the regrets that all approaches achieve are of similar magnitude.
While \ac{fdemd} wins most tasks, ours achieves the best mean rank through more consistent performance.  
Our approach consistently outperforms \ac{dpftsde} and \ac{fmtbo}, except on a few tasks such as that of agent~$6$. 

To determine whether learning a shared representation becomes detrimental under such heterogeneous conditions, we compared \ac{name} with the agent-local baselines L-VBLL and L-GP, neither of which transfers information across agents. If the shared representation induced substantial negative transfer under weakly or non related functions, we would expect \ac{name} to consistently underperform these non-sharing baselines, particularly on weakly related tasks.
Instead, \ac{name} outperforms the local baselines on 14 of the 18 tasks. 
L-VBLL performs only better on Tasks 5 and 8 and L-GP on tasks 6 and 18, which may indicate task-specific interference, but not systematic negative transfer. 
Overall, these results suggest that \ac{name} can exploit shared structure when present while remaining robust to substantial task heterogeneity


These results show that despite lacking explicit mechanisms to identify similar client tasks, 
\ac{name}'s federated surrogate training appears reasonably robust to negative knowledge transfer. 
This is noteworthy as most \ac{fedbo} baselines employ dedicated strategies to identify similar tasks and are hence tailored to this benchmark, while our approach is not and relies on learning them directly.

\definecolor{heat}{HTML}{2B6CB0}   
\sisetup{detect-weight=true,detect-family=true}  
\newcommand{\best}[1]{\textbf{#1}}
\newcommand{\hmcap}{3}             
\newcommand{\hmmax}{60}            
 
\newcommand{\hmcell}[3]{%
  \edef\hmp{heat!\fpeval{round(\hmmax*(1-min(1,ln(#1/#2)/ln(#3))))}}%
  \expandafter\cellcolor\expandafter{\hmp}%
  \ifnum\fpeval{#1==#2}=1 \best{\num{#1}}\else\num{#1}\fi}
 
\newcommand{\hmrow}[7]{%
  \xdef\rowmin{\fpeval{min(#1,#2,#3,#4,#5,#6,#7)}}%
  \xdef\rowspan{\hmcap}
  \hmcell{#1}{\rowmin}{\rowspan} & \hmcell{#2}{\rowmin}{\rowspan} &
  \hmcell{#3}{\rowmin}{\rowspan} & \hmcell{#4}{\rowmin}{\rowspan} &
  \hmcell{#5}{\rowmin}{\rowspan} & \hmcell{#6}{\rowmin}{\rowspan} &
  \hmcell{#7}{\rowmin}{\rowspan}}
 
\newcommand{\hmavgcell}[3]{%
  \edef\hmp{heat!\fpeval{round(\hmmax*(1-(#1-#2)/(#3-#2)))}}%
  \expandafter\cellcolor\expandafter{\hmp}%
  \ifnum\fpeval{#1==#2}=1 \best{#1}\else #1\fi}
\newcommand{\hmavg}[7]{%
  \xdef\rowmin{\fpeval{min(#1,#2,#3,#4,#5,#6,#7)}}%
  \xdef\rowmaxv{\fpeval{max(#1,#2,#3,#4,#5,#6,#7)}}%
  \hmavgcell{#1}{\rowmin}{\rowmaxv} & \hmavgcell{#2}{\rowmin}{\rowmaxv} &
  \hmavgcell{#3}{\rowmin}{\rowmaxv} & \hmavgcell{#4}{\rowmin}{\rowmaxv} &
  \hmavgcell{#5}{\rowmin}{\rowmaxv} & \hmavgcell{#6}{\rowmin}{\rowmaxv} &
  \hmavgcell{#7}{\rowmin}{\rowmaxv}}
 
\begin{table}[t]
\caption{10-dimensional many-tasks results (lower is better). Cell shading is log-scaled relative to the best method per task (darker = better; white = at least $3\times$ worse than the best). The best result per task is in bold.}
\label{tab:experiments:manytask_10d}
\centering
\resizebox{0.9\linewidth}{!}{%
\centering
\small
\setlength{\tabcolsep}{7pt}
\renewcommand{\arraystretch}{1.25}
\begin{tabular}{c *{7}{c}}
\toprule
\textbf{Task} &
\textbf{\ac{name} (ours)} &
\textbf{L-VBLL} & \textbf{L-GP} & \textbf{FMTBO} &
\textbf{DP-FTS-DE} & \textbf{IAF-FBO} & \textbf{FD-EMD} \\
\midrule
1  & \hmrow{9.05E-01}{1.01E+00}{1.04E+00}{1.00E+00}{1.36E+00}{9.44E-01}{7.06E-01} \\
2  & \hmrow{7.33E+01}{1.16E+02}{1.66E+02}{1.08E+02}{3.71E+02}{8.89E+01}{6.68E+01} \\
3  & \hmrow{4.03E+00}{7.58E+00}{1.16E+01}{6.57E+00}{1.54E+01}{6.53E+00}{3.23E+00} \\
4  & \hmrow{7.32E+01}{1.36E+02}{1.76E+02}{1.23E+02}{3.33E+02}{9.18E+01}{6.63E+01} \\
5  & \hmrow{9.28E+00}{7.61E+00}{1.24E+01}{7.70E+00}{1.89E+01}{1.68E+01}{1.84E+01} \\
6  & \hmrow{2.60E+03}{2.67E+03}{2.53E+03}{2.22E+03}{2.72E+03}{2.77E+03}{2.77E+03} \\
7  & \hmrow{7.75E+01}{1.18E+02}{1.73E+02}{1.06E+02}{4.32E+02}{8.80E+01}{6.34E+01} \\
8  & \hmrow{1.76E+02}{1.50E+02}{2.46E+02}{1.78E+02}{1.02E+04}{5.81E+02}{5.89E+01} \\
9  & \hmrow{4.29E+00}{8.06E+00}{1.19E+01}{7.20E+00}{1.53E+01}{6.41E+00}{3.51E+00} \\
10 & \hmrow{1.54E+01}{4.02E+01}{1.89E+02}{9.64E+05}{1.25E+08}{3.45E+05}{5.82E+07} \\
11 & \hmrow{4.19E+00}{7.82E+00}{1.12E+01}{7.29E+00}{1.59E+01}{6.37E+00}{3.36E+00} \\
12 & \hmrow{2.59E+00}{4.85E+00}{8.75E+00}{5.78E+00}{1.17E+01}{3.91E+00}{1.98E+00} \\
13 & \hmrow{1.50E+01}{4.31E+01}{2.16E+02}{1.39E+06}{1.41E+08}{4.06E+05}{5.75E+07} \\
14 & \hmrow{7.65E+01}{1.36E+02}{1.95E+02}{1.33E+02}{4.38E+02}{1.10E+02}{6.76E+01} \\
15 & \hmrow{9.55E-01}{9.86E-01}{1.06E+00}{1.02E+00}{1.41E+00}{9.89E-01}{9.10E-01} \\
16 & \hmrow{2.05E+00}{4.28E+00}{6.60E+00}{4.42E+00}{9.96E+00}{3.01E+00}{1.62E+00} \\
17 & \hmrow{7.23E+01}{1.28E+02}{1.86E+02}{1.14E+02}{4.31E+02}{1.01E+02}{6.35E+01} \\
18 & \hmrow{2.67E+03}{2.68E+03}{2.52E+03}{2.10E+03}{2.87E+03}{2.69E+03}{2.69E+03} \\
\midrule
\textbf{Avg. Rank} & \hmavg{2.11}{3.94}{5.06}{3.83}{6.89}{3.78}{2.39} \\
\bottomrule
\end{tabular}
}
\end{table}

\subsection{Influence of the number of agents on performance}
\label{sec:experiment:ablation}
To evaluate the influence of the number of agents participating in the federated surrogate training, we varied the number of agents with results shown in \cref{fig:experiments:num_agents}.
%
We observe that increasing the number of participating agents improves overall performance, speeding up convergence and reducing variance.
The gains become more pronounced as the problem dimensionality grows.

\ifshowallplotnumberagents
{
\begin{figure}[t]
    \fontsize{\plotfontsize}{\plotfontsize}\selectfont
    \centering

    \includegraphics[width=197.32925pt,height=211.12683pt]{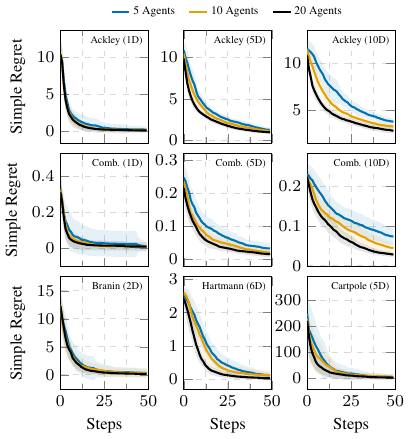}%
\ifshowallplotnumberagents
    \caption{\ac{name} versus number of agents. \capt{More agents increase convergence speed and decrease variance of the outcome.}}
    \label{fig:experiments:num_agents}
\else
    \caption{\ac{name} versus number of agents (extended results). \capt{}}
    \label{fig:experiments:num_agents-extended}
\fi
\end{figure}
}

\else
\begin{wrapfigure}{r}{0.49\textwidth}
    \centering
    \vspace{-1.5em}
    \centering
    \input{plots/ablation_num_agents.tex}
    \vspace{-1em}
\end{wrapfigure}
\fi
These results highlight the effectiveness of leveraging many agents for richer, more diverse exploration of the search space.
With more agents, more data can be used to learn the shared features, directly translating into more accurate shared features and improved optimization performance.
However, this benefit comes at the cost of increased communication overhead, as a larger number of agents must exchange information. 
In practice, the optimal number of agents is therefore problem-dependent and should balance the advantages of enhanced collaboration against the additional communication cost.




\section{Limitations and Future Work}\label{sec:limitations}

\Ac{name} has been shown to significantly accelerate \ac{bo} in \ac{manytaskopt} settings by efficiently exchanging knowledge between agents.
Nevertheless, 
it also has limitations that motivate future research.  


\fakepar{Privacy}
As outlined in \cref{sec:problem-formulation}, \ac{name} preserves privacy in the sense that raw query point-observation pairs are never shared across agents and does not defend against other mechanisms such as gradient inversion attacks~\cite{zhu2019deep, huang2021evaluating, hatamizadeh2023gradient, geng2023improved, zhou2025adversarial}.
However, as our framework is independent of the specific \ac{fl} method, it can in principle support defenses such as gradient perturbation~\cite{zhu2019deep}, input mixing~\cite{huang2020instahide}, or representation perturbation~\cite{sun2021soteria}.
In such scenarios, we expect a trade-off between the strength of the privacy guarantees and optimization performance.

\fakepar{Communication}
Our ablations in Appendix~\ref{sec:appendix:ablation} show that performance degrades when the number of communication rounds per \ac{bo} iteration is reduced, particularly in higher dimensions.
This indicates that \ac{name} currently relies on frequent synchronization to learn effective shared backbones.
In realistic deployments with bandwidth constraints, this requirement may limit scalability and increase optimization time.
Future work could integrate \ac{fl} techniques that reduce communication demands, such as device selection~\cite[Ch.~6.2.2]{Liang_Jin_Ye_Li_2025}, into \ac{name}.

\fakepar{Asynchronous Optimization}
Agents perform their experiments and surrogate training in a fully synchronized manner.
However, many practical scenarios feature heterogeneous experiment durations and computational resources, so stragglers can delay global updates.
This motivates an asynchronous extension of \ac{name}, which is a promising direction for future research.

\section{Conclusion}
This work presented \ac{name}, a novel \ac{fedbo} method that learns a global feature representation to share knowledge between agents during optimization while keeping all raw observations local.
\ac{name} combines a shared \ac{mlp} backbone with agent-specific \ac{vbll} heads and trains the backbone via standard \ac{fl}, enabling knowledge transfer without sharing data.
Across a suite of benchmarks including shifted functions, linear combinations of base functions and a cart-pole controller tuning task, \ac{name} consistently achieves faster convergence and better final performance than existing \ac{fedbo} methods such as \ac{dpftsde} and \ac{fmtbo}, as well as agent-local \ac{bo} with \ac{gp} or \ac{nn} surrogates.
It furthermore matches the performance of a centralized variant that trains on pooled data (Appendix~\ref{sec:appendix:central-baselines}), indicating that aggregating backbone parameters via \ac{fl} is sufficient to capture the benefits of data sharing.
On a many-task benchmark with partially related and unrelated tasks, \ac{name} attains performance comparable to recent clustering-based \ac{fedbo} methods while using a single shared backbone.
Furthermore, late-arriving agents can warm-start their optimization using the shared backbone learned from earlier agents. 
Together, these results demonstrate that shared \ac{nn} representations provide an effective and flexible mechanism for knowledge transfer in \ac{fedbo}, enabling fast convergence without creating overly restrictive assumptions on the setting all while ensuring privacy.


\clearpage

\begin{ack}
This work was supported in part by the German Research Foundation (DFG) through Priority Programme 1914 (project number 315036061), Research Training Group RTG 2236/2 (UnRAVeL), and project number 556142469 (ParDyBO).
Additional support was provided by the German Federal Ministry of Research, Technology and Space (BMFTR) through the Robotics Institute Germany (RIG).

The authors gratefully acknowledge the computing time provided to them at the NHR Center NHR4CES at RWTH Aachen University (project numbers p0021919 and p0027430). This is funded by the Federal Ministry of Research, Technology and Space, and the state governments participating on the basis of the resolutions of the GWK for national high performance computing at universities (\url{www.nhr-verein.de/unsere-partner}).
\end{ack}

\bibliographystyle{plainnat}
\bibliography{references}

\newpage
\appendix
\onecolumn
\renewcommand{\benchmarktitlefontsize}{5}
\section{Surrogate Training Subroutines}
\label{sec:appendix:training}

\Cref{alg:train,alg:aggregate-backbones,alg:train-head} detail the subroutines used in \cref{alg:aprroach_fl_surrogate_training}.
We denote the local and head-only epoch counts by $E$ and $E_{\mathrm{head}}$, respectively.
For a minibatch $\mathcal{B}$, $\ell_i(\headparam{i},\backboneparam{};\mathcal{B})$ denotes the negative variational lower bound in \eqref{eq:variational_lower_bound}, with the data term evaluated on $\mathcal{B}$ and regularization scaled by the full local dataset size $|\dataset{i}|$.
The local optimizer uses Adam with gradient clipping. 
Its state is initialized anew for each call to \textsc{Train} or \textsc{TrainHead}.
The aggregation optimizer instead retains its state across communication rounds within one call to \textsc{FedSurrogateTraining}.

\begin{algorithm}[H]
\caption{Local Surrogate Training on Agent $i$}
\label{alg:train}
\begin{algorithmic}[1]
\footnotesize
    \STATE {\bfseries Input:} Head and backbone parameters $[\headparam{i},\backboneparam{}]$, local dataset $\dataset{i}$, epochs $E$
    \FUNCTION{Train[$[\headparam{i},\backboneparam{}],\dataset{i}$]}
        \STATE Initialize local optimizer state $s$
        \FOR{$e=1$ {\bfseries to} $E$}
            \STATE Form shuffled minibatches from $\dataset{i}$
            \FOR{each minibatch $\mathcal{B}$}
                \STATE $g \gets \nabla_{(\headparam{i},\backboneparam{})}\ell_i(\headparam{i},\backboneparam{};\mathcal{B})$
                \STATE \COMMENT{Jointly update head and backbone}
                \STATE $([\headparam{i},\backboneparam{}],s) \gets \text{LocalOptimizerStep}([\headparam{i},\backboneparam{}],g,s)$
            \ENDFOR
        \ENDFOR
        \STATE {\bfseries return} $\headparam{i},\backboneparam{}$
    \ENDFUNCTION
\end{algorithmic}
\end{algorithm}

\begin{algorithm}[H]
\caption{Backbone Aggregation Using FedOpt~\cite{reddi2021adaptive}}
\label{alg:aggregate-backbones}
\begin{algorithmic}[1]
\footnotesize
    \STATE {\bfseries Input:} Received backbones $\Omega=\{\backboneparami{j}\mid j\in\setagents\}$, previous backbone $\backboneparam{}$, dataset sizes $n_j=|\dataset{j}|$
    \STATE {\bfseries Persistent state:} Aggregation optimizer state $s_{\mathrm{agg}}$, initialized before the first round
    \FUNCTION{AggregateBackbones[$\Omega,\backboneparam{}$]}
        \STATE $\Delta \gets 0$
        \FOR{each agent $j\in\setagents$}
            \STATE $w_j \gets n_j / \sum_{k\in\setagents} n_k$
            \STATE $\Delta \gets \Delta + w_j(\backboneparami{j}-\backboneparam{})$
        \ENDFOR
        \STATE $g_{\mathrm{agg}} \gets -\Delta$ \hfill \COMMENT{Treat the averaged update as a negative gradient}
        \STATE $(\backboneparam{},s_{\mathrm{agg}}) \gets \text{AdamStep}(\backboneparam{},g_{\mathrm{agg}},s_{\mathrm{agg}})$
        \STATE {\bfseries return} $\backboneparam{}$
    \ENDFUNCTION
\end{algorithmic}
\end{algorithm}

\begin{algorithm}[H]
\caption{Local Head Finetuning on Agent $i$}
\label{alg:train-head}
\begin{algorithmic}[1]
\footnotesize
    \STATE {\bfseries Input:} Head and aggregated backbone parameters $[\headparam{i},\backboneparam{}]$, local dataset $\dataset{i}$, epochs $E_{\mathrm{head}}$
    \FUNCTION{TrainHead[$[\headparam{i},\backboneparam{}],\dataset{i}$]}
        \STATE Freeze $\backboneparam{}$ and initialize local optimizer state $s$
        \FOR{$e=1$ {\bfseries to} $E_{\mathrm{head}}$}
            \STATE Form shuffled minibatches from $\dataset{i}$
            \FOR{each minibatch $\mathcal{B}$}
                \STATE $g \gets \nabla_{\headparam{i}}\ell_i(\headparam{i},\backboneparam{};\mathcal{B})$
                \STATE \COMMENT{Adapt the local head to the aggregated features}
                \STATE $(\headparam{i},s) \gets \text{LocalOptimizerStep}(\headparam{i},g,s)$
            \ENDFOR
        \ENDFOR
        \STATE {\bfseries return} $\headparam{i}$
    \ENDFUNCTION
\end{algorithmic}
\end{algorithm}

\section{Benchmark Problems}
\label{sec:appendix:benchmark}

\fakepar{Shift}
The first type of transformation we use to simulate correlation is a simple shift applied to both the input and output of the shared objective, ensuring that each client has a distinct minimum location.
For the shared objectives, we employ standard benchmark functions commonly used in \ac{bo}, namely the Ackley, Branin, and Hartmann functions~\cite{eriksson2019scalable}.

\begin{table}[h]
    \centering
    \small
    \caption{Homogeneous Task Benchmark Specification}
    \vspace{5pt}
    \begin{tabular}{l l r}
        \toprule
        \textbf{Shared Objective} & \textbf{Local Transformation} & \textbf{Input Dimension} \\ 
        \midrule
        Ackley Function & Shift & 1D, 5D, 10D \\
        Branin Function & Shift & 2D \\
        Hartmann Function & Shift & 6D \\
        Combination & Linear Combination & 1D, 5D, 10D \\
        Controller Optimization & Environment Factors & 5D \\
        \bottomrule
    \end{tabular}
    \label{tab:homogeneous_tasks}
\end{table}

\fakepar{Linear Combination}
As motivated in \cref{sec:problem-formulation}, some \ac{manytaskopt} problems can be modeled as a linear combination of shared components, weighted by user-specific preferences.
Therefore, we design a benchmark problem to reflect this relationship by defining each agent's objective as a unique combination of five base benchmark functions.
The objective function for agent $k$ is defined as:
\begin{equation}
\label{eq:results_linear_combination_function}
    \text{Combination}^{(k)}(\mathbf{x}) = 
        \mathbf{a}^{(k)}
        \cdot
        \begin{bmatrix}
            \text{Ackley}(x) \\
            \text{Cosine}(x + 1) \\
            \text{Rastrigin}(x - 8) \\
            \text{Griewank}(x + 4) \\
            \text{Sphere}(x - 3)
        \end{bmatrix}
\end{equation}
where $\mathbf{a}^{(k)}$ is the agent-specific row vector of coefficients.
The coefficients in $\mathbf{a}^{(k)}$ are randomized for each agent. This ensures that each agent has a differently shaped objective function, modeling unique preferences.
Additionally, we apply fixed shifts to the inputs of the base functions. 
This prevents trivial alignment of the functions and allows for different optimal locations to emerge depending on the agent's coefficient vector.
The output of each base benchmark function is scaled to the range $[0, 1]$. This normalization ensures that all components contribute equitably to the final combination, regardless of their original output scales.

\fakepar{Controller Optimization}
The final type of \ac{manytaskopt} problem we include is an application scenario, where we tune controllers operating under different environmental conditions. 
Specifically, each client tunes a \ac{lqr} controller for the classical cart-pole problem \cite{cartpole}.
The cart-pole system consists of a pole attached to a cart that moves along a track.
The objective is to stabilize the system by applying horizontal forces to the cart while keeping the pole upright.
The system's dynamics are nonlinear and are typically described by a set of coupled differential equations involving the cart's position and velocity, as well as the pole's angle and angular velocity.

To construct the controller, a linearized model around the upright equilibrium is used, from which an \ac{lqr} feedback law is obtained that minimizes
\begin{equation}
J = \int_{0}^{\infty} (\mathbf{x}^T(t) Q \mathbf{x}(t) + \mathbf{u}^T(t) R \mathbf{u}(t)) \,dt
\end{equation}
where $\mathbf{x}(t)$ denotes the system state, $\mathbf{u}(t)$ the control input, and $Q$ and $R$ are positive semi-definite weighting matrices that balance the trade-off between state error and control effort.
Rather than directly optimizing the quadratic \ac{lqr} cost, each agent treats the weighting matrices $Q$ and $R$ as tunable parameters that define a feedback controller through the solution of an algebraic Riccati equation as in \citet{marco2016automatic}.
In our setup, $Q$ is restricted to be diagonal and $R$ is scalar, so that each agent’s parameter vector consists of the four diagonal entries of $Q$ and the single entry of $R$.

We randomize the environmental factors for each agent, including parameters such as the mass of the cart or the length of the pole.
This variation causes the behavior of the systems to differ, while their underlying cart-pole dynamics remain the same.

The objective function $f^{(k)}$ for agent $k$ is defined as the negative average number of time steps for which the pole remains upright and within a certain range of the track.
For a given set of $Q$ and $R$ parameters, we simulate the agent's specific cart-pole system for ten episodes, each running for a maximum of $500$ time steps.
The function value is the negative average number of steps the pole remained stable across these episodes. This formulation turns the stabilization task into a minimization problem.

\begin{table}[t]
    \centering
    \small
    \caption{Heterogeneous Task Benchmark Specification based on \cite{da2017evolutionary}}
    \vspace{5pt}
    \begin{tabular}{l l l r}
    \toprule
    Category & Task & Degree of intersection & $R_s$ \\
    \midrule
    \multirow{2}{*}{CI+HS} & Griewank ($T_1$) & \multirow{2}{*}{Complete intersection} & \multirow{2}{*}{1.000} \\
     & Rastrigin ($T_2$) &  & \\
    \midrule
    \multirow{2}{*}{CI+MS} & Ackley ($T_3$) & \multirow{2}{*}{Complete intersection} & \multirow{2}{*}{0.2261} \\
     & Rastrigin ($T_4$) &  & \\
    \midrule
    \multirow{2}{*}{CI+LS} & Ackley ($T_5$) & \multirow{2}{*}{Complete intersection} & \multirow{2}{*}{0.0002} \\
     & Schwefel ($T_6$) &  & \\
    \midrule
    \multirow{2}{*}{PI+HS} & Rastrigin ($T_7$) & \multirow{2}{*}{Partial intersection} & \multirow{2}{*}{0.8670} \\
     & Sphere ($T_8$) &  & \\
    \midrule
    \multirow{2}{*}{PI+MS} & Ackley ($T_9$) & \multirow{2}{*}{Partial intersection} & \multirow{2}{*}{0.2154} \\
     & Rosenbrock ($T_{10}$) &  & \\
    \midrule
    \multirow{2}{*}{PI+LS} & Ackley ($T_{11}$) & \multirow{2}{*}{Partial intersection} & \multirow{2}{*}{0.0725} \\
     & Weierstrass ($T_{12}$) &  & \\
    \midrule
    \multirow{2}{*}{NI+HS} & Rosenbrock ($T_{13}$) & \multirow{2}{*}{No intersection} & \multirow{2}{*}{0.9434} \\
     & Rastrigin ($T_{14}$) &  & \\
    \midrule
    \multirow{2}{*}{NI+MS} & Griewank ($T_{15}$) & \multirow{2}{*}{No intersection} & \multirow{2}{*}{0.3669} \\
     & Weierstrass ($T_{16}$) &  & \\
    \midrule
    \multirow{2}{*}{NI+LS} & Rastrigin ($T_{17}$) & \multirow{2}{*}{No intersection} & \multirow{2}{*}{0.0016} \\
     & Schwefel ($T_{18}$) &  & \\
    \bottomrule
    \end{tabular}
    \label{tab:heterogeneous_tasks}
\end{table}


\section{Hyperparameter}
\label{sec:appendix:hyperparameters}

Unless otherwise specified, we use a fully connected backbone with 4 hidden layers and 64 neurons per layer (\cref{tab:fl_hyperparams}). 
For the model head, we employ a single output layer that depends on the chosen type of neural network, with ReLU as the activation function for all neurons. 
The \ac{vbll} models use the variational lower bound for regression defined in \cref{eq:variational_lower_bound}. 
For the \ac{gp} baseline, we employ a Matérn kernel as well as TS as an acquisition function using a candidate set as described in \citet[Appendix~E]{eriksson2019scalable}.

Moreover, similar to \citet[Appendix~C.1]{brunzema2025bovbll} this appendix also evaluates a second variant of our approach that employs \ac{mlp} networks as surrogates instead of a \ac{vbll}, \ie{} it learns the Thompson sample. 
We refer to this variant as \ac{name}~(MLP) for the \ac{fedbo} implementation, L-MLP for the agent-local implementation and C-MLP for the central variant (cf. \cref{sec:appendix:central-baselines}).
These \ac{mlp}-based surrogates are trained using a Mean Squared Error loss.

\begin{table}[h]
\centering
\small
\caption{Hyperparameter of neural network surrogate models.}
\vspace{5pt}
\label{tab:fl_hyperparams}
\begin{tabular}{l c c c c c}
\toprule
\textbf{Approach} & \textbf{Local LR} & \textbf{Agg. LR} & \textbf{Features} & \textbf{Backbone Layers} & \textbf{Head Layers} \\
\midrule
FL-MLP & 1e-4 & 8e-3 & 64 & 4 & 1 \\
FL-VBLL & 1e-3 & 8e-3 & 64 & 4 & 1 \\
\midrule
L-MLP & 1e-4 & / & 64 & / & 5 \\
L-VBLL & 1e-4 & / & 64 & / & 5 \\
\bottomrule
\end{tabular}
\end{table}

\begin{table}[h]
\centering
\small
\caption{\ac{vbll} Hyperparameters.}
\vspace{5pt}
\label{tab:vbbl_hyperparams}
\begin{tabular}{l c c c}
\toprule
\textbf{Approach} & \textbf{Prior Scale} & \textbf{Wishart Scale} & \textbf{Degrees of Freedom}  \\
\midrule
FL-VBLL &  10.0 & 1.0 & 25.0 \\
\midrule
L-VBLL & 10.0 & 0.01 & 1.0 \\
\midrule
C-VBLL & 10.0 & 1.0 & 25.0 \\
\midrule
FL-MLH-VBLL & 10.0 & 1.0 & 25.0 \\
\bottomrule
\end{tabular}
\end{table}
 {
\begin{figure}[t]
    \fontsize{\plotfontsize}{\plotfontsize}\selectfont
    \centering

    \includegraphics[width=205.32925pt,height=181.91779pt]{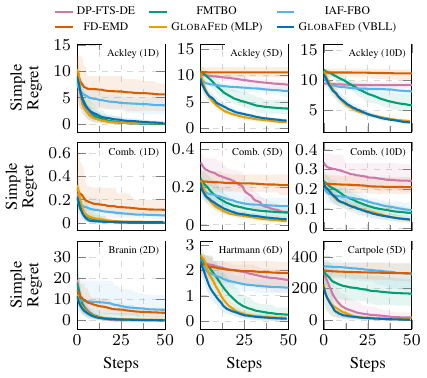}\acused{name}\acused{dpftsde}\acused{fmtbo}\acused{iaffbo}\acused{fdemd}\acused{vbll}\acused{mlp}%
\caption{Performance of \ac{fedbo} baselines in comparison to the \ac{vbll} and \ac{mlp} variants of \ac{name}. \capt{All approaches perform similar in low dimensions, except \ac{iaffbo} and \ac{fdemd}, which do not converge to the optimum. \ac{name} converges much faster and to lower final regrets for high-dimensional problems than all \ac{fedbo} baselines.}}
    \label{fig:experiments:homogeneous-fedbo-extended}
\end{figure}
}

\section{Extended Results}\label{sec:appendix:extended-results}
\newif\ifcentralbenchmarkpaired
\ifshowallplotlocal
    \ifshowallplotwarmstart\else
        \begin{figure}[t]
            \centering
            \begin{minipage}[t]{0.48\textwidth}
                \input{plots/homogeneous_benchmark_fedbo_warmstart_extended.tex}
            \end{minipage}
        \end{figure}
    \fi
\else
    \begin{figure}[t]
        \centering
        \begin{minipage}[t]{0.48\textwidth}
            \vspace{0pt}
            \input{plots/homogeneous_benchmark_local_extended.tex}
        \end{minipage}
        \ifshowallplotwarmstart
            \global\centralbenchmarkpairedtrue
            \hfill
            \begin{minipage}[t]{0.48\textwidth}
                \vspace{0pt}
                \input{plots/homogeneous_benchmark_central.tex}
            \end{minipage}
        \else
            \hfill
            \begin{minipage}[t]{0.48\textwidth}
                \input{plots/homogeneous_benchmark_fedbo_warmstart_extended.tex}
            \end{minipage}
        \fi
    \end{figure}
\fi

\Cref{fig:experiments:homogeneous-fedbo-extended} shows the comparison of \ac{name} with other \ac{fedbo} baselines for all objective functions.
\ifshowallplotlocal\else
\Cref{fig:experiments:homogeneous-local-extended} shows the comparison with agent-local baselines for all objective functions.
\fi
\ifshowallplotwarmstart\else
\Cref{fig:experiments:homogeneous-fedbo-warmstart-extended} shows the warmstarting comparison for all objective functions.
\fi
The effects discussed in \cref{sec:experiments:homogeneous-benchmark}, namely similar behavior of all approaches except \ac{iaffbo} and \ac{fdemd} for low dimensions  and significantly faster optimization via \ac{name} than baselines for higher dimension, remain consistent among all objectives.

Both the \ac{mlp} and \ac{vbll} variants of \ac{name} perform similarly.
We believe this parity is mainly due to two aspects. 
First, we utilize Thompson sampling in this pipeline, which we employed specifically to maintain consistency with the original \ac{vbll} paper's method~\cite{brunzema2025bovbll}. The authors of \cite{brunzema2025bovbll} noted a slight advantage for \acp{vbll} over standard MLPs.
And second, we hypothesize that training for more epochs with early stopping could further improve \ac{vbll} performance as demonstrated in prior work on \ac{vbll} surrogate models~\cite{brunzema2025bovbll}. 

\ifshowallplotlocal\else
Only the \ac{nn}-based agent-local approaches perform poorly for low-dimensional problems.
We attribute the poor performance to suboptimal initialization that can trap the \ac{mlp} surrogate, causing repeated sampling at the search-space boundaries.\footnote{Note that the architecture obtained from the 10D Ackley task differs from the one used in \citet{brunzema2025bovbll}.}
Our federated approach mitigates this issue: by sharing the model structure across agents, information from other clients can correct poorly initialized or stuck surrogates, resulting in more robust and stable optimization behavior.
\fi

\section{Training of Central Baselines}
\label{sec:appendix:central-baselines}

To examine whether our privacy-preserving knowledge transfer incurs any performance loss compared to centralized training with full data access, we designed a centralized version of our approach.
In this setting, a shared backbone is optimized using a global loss defined over the data of all agents.
Let $(X_{i}, Y_{i})$ be the batch of input and target data for agent $i$, and $\mathcal{L}$ be the loss function. We define the global loss as:
\begin{equation}
\label{eq:centralized_loss}
\mathcal{L}_{\text{global}} = \frac{1}{K}\sum\limits_{k=1}^K \mathcal{L}(X_{i}, Y_{i}, (\backboneparam, \headparam{i})).
\end{equation}
Training follows the same procedure as the federated surrogate training illustrated in \cref{sec:method:surrogate}, except that the backbone is trained jointly using the global loss instead of via \ac{fl}.
Note that especially for the \ac{vbll} baseline, formulating a joint optimization objective over multiple heads is non-trivial as they may converge at different rates and it is an interesting direction for future work to prevent overfitting of individual heads.
However, the results in Table~\ref{tab:experiments:manytask_10d} show that a simple aggregation as in \eqref{eq:centralized_loss} already yields good performance.

\ifcentralbenchmarkpaired\else
\begin{figure}[t]
    \centering
    \begin{minipage}[t]{0.48\textwidth}
        \input{plots/homogeneous_benchmark_central.tex}
    \end{minipage}
\end{figure}
\fi

\Cref{fig:experiments:homogeneous-central} compares our approach with its centralized variants to assess whether knowledge transfer via model aggregation is as effective as training a centralized backbone on the raw data of all agents. 
Our federated method matches the performance of the centralized approach, with small problem-dependent differences where either variant may perform slightly better. 
We therefore conclude that the privacy-preserving nature of our method does not degrade performance. 
These results suggest that shared feature representations, whether learned centrally or federated, provide a powerful and robust mechanism for knowledge transfer across agents.

\section{Ablations}\label{sec:appendix:ablation}
\ifshowallplotnumberagents\else
{
\begin{figure}[t]
    \fontsize{\plotfontsize}{\plotfontsize}\selectfont
    \centering

    \includegraphics[width=197.32925pt,height=211.12683pt]{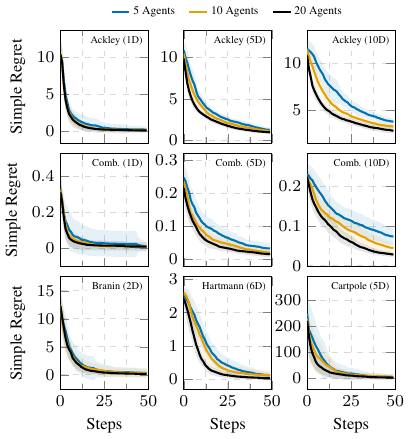}%
\ifshowallplotnumberagents
    \caption{\ac{name} versus number of agents. \capt{More agents increase convergence speed and decrease variance of the outcome.}}
    \label{fig:experiments:num_agents}
\else
    \caption{\ac{name} versus number of agents (extended results). \capt{}}
    \label{fig:experiments:num_agents-extended}
\fi
\end{figure}
}

\subsection{Number of Agents}
\Cref{fig:experiments:num_agents-extended} presents the results of our study regarding the number of agents for all objectives functions showing consistent behavior across all functions, \ie more agents improve convergence and reduce variance especially for higher-dimensional problems.

\fi

\subsection{Number of Communication Rounds}\label{sec:appendix:ablation:num-com-rounds}
For our experiments, we use a high number of communication rounds $R$ per optimization step to ensure optimal performance.
However, for real-world applications, this introduces high communication costs and synchronization overhead.
Therefore, we tested how reducing the number of communication rounds while keeping the overall number of training epochs the same affects performance. Specifically, we evaluated our approach for $R=25$, $R=50$, and $R=100$ communication rounds.

As \cref{fig:experiments:num-com-rounds} illustrates, the highest number of rounds achieves the best overall performance.
A possible explanation for this behavior is that more frequent communication enables more direct and consistent knowledge transfer, thereby improving the quality of the surrogate model approximation.
Furthermore, each communication round includes an aggregation step, which serves as an additional update step during training.
Configurations with fewer rounds inherently perform this update step less often.
The performance impact of reducing communication rounds is less prominent for simpler problems.
While fewer rounds still achieve comparable performance on the linear combination and controller optimization tasks, they result in significantly worse performance and higher variance on the shifted benchmark problems.
This suggests that the shifted problems are inherently more complex to model with our surrogate, and therefore require more frequent communication to learn a sufficiently expressive shared feature representation.

We want to note that our method does incur higher communication costs than the considered baselines. Existing approaches typically require only one or a small number of model exchanges per newly evaluated objective point, whereas our decentralized knowledge-sharing mechanism may involve hundreds of communication rounds.

This additional overhead is a deliberate trade-off: more frequent communication enables agents to continuously refine the shared representation and thereby improve sample efficiency without requiring additional, potentially expensive objective evaluations. 
We therefore expect our method to be most attractive in settings where objective evaluations are substantially more costly than communication.


\subsection{Feature Dimensionality}
The number of features per layer determines the total number of parameters each agent must transmit, thereby directly affecting the size of the messages exchanged during communication.
Consequently, it has a direct impact on the communication cost of the entire knowledge transfer process.

As shown in \cref{fig:experiments:feature-dim}, the optimal number of features depends on the complexity of the underlying problem.
While a small number of features is sufficient to model low-dimensional problems, more complex problems require larger feature representations to capture intricate relationships between tasks.
Reducing the number of features to $16$ per layer already incurs a loss in performance for the $2$-dimensional shifted Branin benchmark. 
As the dimension of the problem increases, so does the loss of performance for using a low number of features.

A higher number of features per layer can directly improve the performance of our approach, but also increases its communication costs.
However, the performance improvement might only be minor if the complexity of the problem does not increase in the same way.
Consequently, it is essential to strike a balance between model expressiveness and communication efficiency.
As a general guideline, we suggest tuning the surrogate model on a fast-to-evaluate test problem of similar dimensionality to the real application to determine optimal parameters.


\subsection{Backbone Depth}
In combination with the number of hidden features, the backbone depth determines the message size of the federated surrogate training.
Contrary to previous results, where increasing a hyperparameter, such as the number of rounds or features, usually improved performance, \cref{fig:experiments:beackbone-depth} shows that this does not necessarily hold for the number of backbone layers. 
In our experiments, a backbone with only two layers performed the best overall.

This observation may be closely related to the amount of local data available. 
Since agents optimize objective functions that are expensive to evaluate, their local datasets are small. While deeper backbones can theoretically capture more complex relationships, they also require more data to train effectively.
In this low-data context, the smallest backbone appears to produce better shared features, suggesting more stable training.
Smaller models are also less prone to overfitting and can learn more robust features from limited data.
While a network that is too shallow may lack the expressive power to model complex relationships, our findings indicate that a small backbone depth was sufficient for the specific relations we simulated, even for high-dimensional problems.

\subsection{Head Depth}
The number of head layers does not influence the communication cost of our approach, since the head parameters are not transmitted.
However, it remains an important hyperparameter to investigate, as it determines how efficiently each agent can transform and adapt the shared backbone features to its local problem.

A single output layer includes no activation functions and can therefore only combine the shared features as a linear combination. Our previous results on the high-dimensional heterogeneous task benchmark already indicated that more complex adaptation is especially important for heterogeneous tasks that may have only limited similarities.
The effect of the number of head layers is presented in \cref{fig:experiments:head-depth}.

For low-dimensional problems, a single output layer is typically sufficient and can even be advantageous. While adding one additional head layer performs comparably in these cases, introducing too many layers can degrade performance and increase variance, as observed for configurations with three head layers.
A possible explanation for this behavior could be that introducing additional layers reduces the direct influence of the shared features and can result in overfitting on the local task. As we have illustrated in \cref{fig:experiments:homogeneous-local}, shared knowledge is vital for neural network surrogates in low dimensions to prevent the agent from becoming stuck in sampling the boundaries of the search space.
In contrast, for high-dimensional problems, adding more head layers proves beneficial, as it allows for a more flexible adaptation of the shared features to the increased problem complexity.

\begin{figure*}[th!]
    \centering
    \begin{minipage}[t]{0.48\textwidth}
        \input{plots/ablation_number_communication_rounds.tex}
    \end{minipage}
    \hfill
    \begin{minipage}[t]{0.48\textwidth}
        \input{plots/ablation_number_feature_dim.tex}
    \end{minipage} \\
    \begin{minipage}[t]{0.48\textwidth}
        \input{plots/ablation_backbone_depth.tex}
    \end{minipage}
    \hfill
    \begin{minipage}[t]{0.48\textwidth}
        \input{plots/ablation_head_depth.tex}
    \end{minipage}%
\end{figure*}

Interestingly, across all benchmark problems, models with a single output layer often achieve better performance during the early iterations. 
This may be because only linearly combining the shared features allows them to exert a stronger influence on the predictions. This increased reliance on shared knowledge might, in turn, encourage exploration using the shared features, which proves to be advantageous in early iterations.
As a consequence, it might be beneficial to employ an adaptive partitioning of the surrogate model for optimal performance. Early iterations could strongly leverage shared knowledge by employing a single output layer, while later iterations could switch to a complex head to adapt to the local problem as more local observations become available.


\end{document}